\documentclass{article}
\usepackage[utf8]{inputenc}
\usepackage{a4wide}
\usepackage{xcolor}
\usepackage{colortbl}
\usepackage{graphicx}
\usepackage{longtable}
\usepackage{multirow}
\usepackage{listings}
\usepackage{booktabs}
\usepackage{url}
\usepackage{natbib}

\begin{document}

\title{Analysis of Systems' Performance in Natural Language Processing Competitions}
\author{Sergio Nava-Muñoz$^{1}$ and Mario Graff$^{2}$ and Hugo Jair Escalante$^{3}$}

\date{\small{$^{1}$ CIMAT, Aguascalientes, Mexico} ~\\
\small{$^{2}$ INFOTEC, Aguascalientes, Mexico}~\\
\small{$^{3}$ INAOE, Puebla, Mexico}~\\
nava@cimat.mx, mario.graff@infotec.mx, hugojair@inaoep.mx~\\
This work has been accepted for publication in Pattern Recognition Letters. ~\\ \url{https://doi.org/10.1016/j.patrec.2024.03.010}}

\maketitle

\begin{abstract}
    Collaborative competitions have gained popularity in the scientific and technological fields. These competitions involve defining tasks, selecting evaluation scores, and devising result verification methods. In the standard scenario, participants receive a training set and are expected to provide a solution for a held-out dataset kept by organizers. An essential challenge for organizers arises when comparing algorithms' performance, assessing multiple participants, and ranking them. Statistical tools are often used for this purpose; however, traditional statistical methods often fail to capture decisive differences between systems' performance. This manuscript describes an evaluation methodology for statistically analyzing competition results and competition. The methodology is designed to be universally applicable; however, it is illustrated using eight natural language competitions as case studies involving classification and regression problems. The proposed methodology offers several advantages, including off-the-shell comparisons with correction mechanisms and the inclusion of confidence intervals. Furthermore, we introduce metrics that allow organizers to assess the difficulty of competitions. Our analysis shows the potential usefulness of our methodology for effectively evaluating competition results.   
\end{abstract}

\section{Introduction}
\label{sec:introduction}

A challenge is a collaborative competition targeting a specific problem under a controlled scenario and using well-defined evaluation protocols. These competitions span various scientific and technological fields, encompassing fundamental and applied questions in machine learning~\citep{JMLR:v24:21-1436}. Commonly, competitions become established benchmarks that push the state-of-the-art; for example, the ImageNet~\citep{imagenet} and VOC~\citep{voc} competitions.

Introduced by Jeff Howe in Wired magazine in 2006, the concept of crowdsourcing harnesses the power of collective intelligence to accomplish tasks, transcending traditional boundaries to include fields as varied as marketing, astronomy, and genetics. Its application is particularly transformative in the scientific arena, where it taps into the community's collective wisdom to solve complex scientific challenges. The emergence of collaborative competitions, known as ``challenges'', marks a significant trend in science and technology, where diverse participants' collective ingenuity and innovative capacities are directed toward groundbreaking discoveries and solutions.

Within these challenges, participants receive a dataset and a specific question to be answered. The challenge organizers kept hidden the ground truth or gold standard data, which is used as an unbiased benchmark for evaluating participants' techniques. Participants submit their solutions for their evaluation with the gold standard, enabling the identification of the most effective method for solving the given problem while also providing an objective assessment of their approaches. To effectively organize these challenges, it is crucial to establish precise scoring mechanisms for evaluating solutions. 

A challenge task involves comparing the performance of algorithms where certain constraints come into play. These include assessing multiple participants (algorithms, methods, or ensembles), selecting appropriate performance metrics, working with a fixed dataset size, and limiting the number of submissions per participant. Traditional statistical methods for inferring the significance of a particular performance metric are challenging to apply due to the absence of multiple datasets or extensive submissions. The focus of this research is to show how organizers can make effective comparisons under these constraints.

Specifically, our methodology is designed to complement the standard \emph{winner} selection process, which in most cases relies solely on score ranking, by introducing well-established statistical tools and graphics to facilitate analysis. The evaluation methodology is implemented in a ready-to-use tool (namely, CompStats\footnote{http://compstats.readthedocs.org}) that any competition organizer can use. The benefits of the proposed approach are illustrated using various challenge datasets associated with Natural Language Processing (NLP) tasks. These challenges comprise diverse scores used to rank classifiers and regressors. The methodology assesses performance disparities, ascertains whether the observed variations in test data performance hold statistical significance, and performs a comparative analysis of a broad spectrum of NLP competitions. Also, we evaluated the degree of competitiveness among these challenges, considering their potential for improvement in future iterations. Furthermore, we examined whether the designated winner's advantage was discernible. Our experimental evaluation reveals the importance of performing an adequate analysis of results. We foresee that our analysis tool can become a reference methodology for comparing results in collaborative competitions. 

This work builds upon our previous work where standard statistical analysis was adopted to evaluate a single NLP competition~\citep{10.1007/978-3-031-33783-3_9}. This paper goes several steps further: (i) we augmented the statistical tools available for analysis, including correction tests; (ii) graphical analysis and measures for determining the difficulty of a competition are introduced; (iii) showcasing the application of our methodology on eight NLP competitions as examples of its potential.

The remainder of this manuscript is organized as follows: Collaborative competitions are described in Section \ref{sec:competitions}. Section \ref{sec:dataset} explains the dataset used to illustrate this proposal. Our proposed solutions are detailed in Section \ref{sec:approach}. The comparison of competitions is presented in Section \ref{sec:CompetitionsComparison}, where their results are evaluated. Finally, conclusions are provided in Section \ref{sec:conclusions}.

\section{Competitions considered in our study}
\label{sec:competitions}

Our proposed evaluation methodology is versatile and designed to apply universally to any challenge. We selected several NLP challenges as case studies to demonstrate their effectiveness and applicability. These competitions, encompassing various tasks and evaluation metrics, provide diverse scenarios that showcase the strength of our approach. Below, we briefly describe each competition, highlighting the specific tasks and metrics for ranking participant systems.

\textbf{MEX-A3T 2019} \citep{Aragon2019} consists of two tracks. The first one, \textbf{Author Profiling}, aims to determine the gender, occupation, and place of residence of Twitter users in Mexico based on their tweets. It incorporates text and images as information sources to assess their relevance and complementarity in user profiling. Evaluation for this track is conducted using the macro-averaged F1 score. The second track, \textbf{Aggressiveness Detection}, focuses on identifying aggressive tweets in Mexican Spanish. The evaluation is based on the F1 score in the ``aggressiveness'' class.

\textbf{TASS 2020} \citep{Garcia-Vegaa2020} consists of two tracks, namely \textbf{General Polarity at Three Levels} and \textbf{Emotion Detection}; however, in this analysis, we solely focused on the former. The objective is to evaluate polarity classification systems for tweets written in Spanish and their different variants. Participant systems in this competition were ranked based on the macro-averaged F1 score.

The \textbf{VaxxStance 2021} challenge \citep{Agerri2021} aims to determine the stance expressed on the highly controversial topic of the anti-vaxxers movement in two languages: Basque and Spanish. The primary objective is identifying whether a given tweet conveys an ``against'', ``favor'', or ``neutral'' (none) stance regarding this predefined topic.\footnote{\textbf{Open track} and \textbf{Zero-shot track} were not considered because  of too limited participation}. The competition introduced specific participation categories for Basque and Spanish, referred to as the \textbf{Close Track}. Within this track, participant systems are presented with two evaluation choices: \textbf{Textual}, enabling them to work exclusively with the provided tweets in the target language during development, and \textbf{Contextual}, which permits the utilization of supplementary Twitter-related data, including user-based features, friend connections, and retweet information. The Macro-averaged F1 score was also utilized for these subtasks. Nevertheless, it was exclusively applied to two classes, ``favor'' and ``against,'' despite the presence of the ``none'' class in the dataset.

\textbf{EXIST 2021} \citep{Rodriguez-Sanchez2021}: Sexism Identification in Social Networks. Participant systems classify tweets and gab posts (in English and Spanish) according to the following two tasks. The \textbf{Sexism Identification} task aims to determine whether a given text is sexist. Evaluation for this track is done using the accuracy. The \textbf{Sexism Categorization} task uses only sexist texts; it categorizes the message based on the type of sexism. The macro-averaged F1 score ranks the participant systems.

\textbf{DETOXIS 2021} \citep{Taule2021} (DEtection of TOxicity in comments In Spanish) primarily aims to identify toxicity in Spanish comments posted in response to online news articles related to immigration. Specifically focusing on the \textbf{Toxicity Detection} task, it involves classifying comment content as toxic or non-toxic, with participant systems' performance ranked based on F1 scores.

\textbf{MeOffendEs 2021} \citep{Plaza-del-Arco2021} contributes to the progress of research in identifying offensive language across various Spanish-language variations.\footnote{This challenge consists of four subtasks, but only subtask three was used.} The subtask analyzed involves \textbf{Mexican Spanish non-contextual binary classification}, where participant systems categorize tweets from the OffendMEX corpus as offensive or non-offensive. The evaluation is based on the F1 score of the offensive class.

\textbf{REST-MEX 2021} \citep{Alvarez-Carmona2021} encompasses two objectives: a \textbf{Recommendation System} and \textbf{Sentiment Analysis} utilizing text data from Mexican tourist destinations. The Recommendation System task involves forecasting the level of satisfaction a tourist might experience when suggesting a destination in Nayarit, Mexico, based on the places they visited and their feedback. Conversely, the Sentiment Analysis task determines the sentiment expressed in a review provided by a tourist who visited the most iconic locations in Guanajuato, Mexico. This competition ranked the participant systems using the metric \textit{mean square error} (MAE).

\textbf{REST-MEX 2022} \citep{Alvarez-Carmona2022} has three tasks: \textbf{Recommendation System} (not analyzed), \textbf{Sentiment Analysis}, and \textbf{Epidemiological Semaphore}. The Sentiment Analysis one involves classifying sentiments in tourist reviews about Mexican destinations, ranging from 1 (most negative) to 5 (most positive), with attractiveness assessment classes: Attractive, Hotel, and Restaurant, evaluated using the $measure_S$ metric \cite{Alvarez-Carmona2022}. Based on COVID news, the Epidemiological Semaphore task predicts the Mexican Epidemiological Semaphore. It employs a four-color system (red, orange, yellow, green) with varying restrictions across Mexican states. It is assessed using the $measure_C$ metric \cite{Alvarez-Carmona2022}.

\textbf{PAR-MEX 2022} \citep{Bel-Enguix2022} (\textbf{Paraphrase Identification} In Mexican Spanish) consists in determining whether a pair represents a paraphrase relationship, i.e., classifying them as either paraphrases or non-paraphrases; the competition utilized the F1 score as the ranking metric for participant systems. 

In all the analyzed competitions, the datasets include all participant systems, except in EXIST, where the top 10 for individual languages (English and Spanish) were included. On the other hand, EXIST, TASS, DETOXIS, PAR-MEX, and MeOffendEs analyzed the best runs from each participant system. The dataset comprises unique participant systems, with no repetitions of participant systems with different runs, unlike the other competitions where all submitted runs were included. Another additional consideration is that REST-MEX 2021 and EXIST included a {\em baseline,} while MeOffendEs included two. Finally, in REST-MEX 2022, the majority class was included. In this way, when we refer to competitors, we may be referring to different runs by the competitors or even a baseline or the majority class. All the metrics used in the analyzed competitions are of the higher is better, except for MAE, where lower is better.

\section{VaxxStance 2021 - Multiclass Textual Classification }
\label{sec:dataset}

In order to provide insight into the information analyzed in a competition, in this section, we describe the VaxxStance dataset; particularly, it describes subtask 1. The dataset comprises six variables corresponding to the predictions of three teams and the gold standard. Two teams contributed two runs each, derived from the same model under different configurations, while one provided only one run. The gold standard for the Basque dataset consists of 312 tweets: 85 in favor, 135 neutral, and 92 against. Meanwhile, the Spanish dataset has 694 tweets, with 359 in favor, 195 neutral, and 140 against.

The evaluation of the systems was based on the F1 macro-average score for two classes, FAVOR and AGAINST, despite the presence of a NONE class in the test data.

\begin{equation}
    F1_{avg} = \frac{F1_{favor} + F1_{against}}{2}
\end{equation}

Table \ref{tab:metricas} summarizes the results using the F1 macro-average score for Basque and Spanish. As can be seen, the highest score for Basque is $0.5734$ achieved by $WordUp$ in their first run, followed by themselves with the second run, which scored $0.5465$. In third place is $MultiAztertest$ with a score of $0.5024$ in their first run. For Spanish, the three best scores were $WordUp$ in the second and first runs, and $MultiAztertest$ in their first run with values of $0.8092$, $0.7906$, and $0.7410$, respectively.

\begin{table}[!h]
\caption{\label{tab:metricas}Results for VaxxStance Close track - contextual.}
\scriptsize
\centering
\begin{tabular}[t]{lrlll}
\toprule
  & Basque  & & & Spanish \\
\midrule
\cellcolor{gray!6}{Gold\_Standard} & \cellcolor{gray!6}{1.0000}  & & \cellcolor{gray!6}{Gold\_Standard} &\cellcolor{gray!6}{1.0000}\\
WordUp.01 & 0.5734  & & WordUp.02 &0.8092\\
\cellcolor{gray!6}{WordUp.02} & \cellcolor{gray!6}{0.5465}  & & \cellcolor{gray!6}{WordUp.01} &\cellcolor{gray!6}{0.7906}\\
MultiAztertest.01 & 0.5024  & & MultiAztertest.01 &0.7410\\
\cellcolor{gray!6}{SQYQP.01} & \cellcolor{gray!6}{0.4256}  & & \cellcolor{gray!6}{SQYQP.01} &\cellcolor{gray!6}{0.6738}\\
MultiAztertest.02 & 0.3428  & & MultiAztertest.02 &0.6404\\
\bottomrule
\end{tabular}
\end{table}

\section{Proposed methodology}
\label{sec:approach}

The research aim is to introduce tools that facilitate the comparison of results among different competitors, as those shown in Table \ref{tab:metricas} for Basque and Spanish. In the existing literature, studies address the issue of comparing classification algorithms; however, these primarily address aspects other than the competition framework. For instance, \cite{Dietterich1998} reviews five closely related statistical tests for assessing whether one learning algorithm outperforms another in specific learning tasks. Nevertheless, these tests require access to the underlying algorithm. At the same time, in a competition scenario, there is only access to the predictions, not the algorithms themselves. On a different note, \cite{JMLR:v7:demsar06a} focuses on the Statistical Comparisons of Classifiers across Multiple Data Sets. However, the scenario involves just one dataset. Dem{\v{s}}ar presents several non-parametric methods and guidelines for conducting a proper analysis when comparing sets of classifiers. \cite{Garcia2008} address a problem akin to Dem{\v{s}}ar's but concentrate on pairwise comparisons, precisely statistical procedures for comparing $c \times c$ classifiers. However, their focus still revolves around scenarios with multiple datasets employing the same classifiers.

\subsection{Bootstrap}
\label{sec:bootstrap}

The term {\em bootstrapping} in statistics refers to the practice of making conclusions about the sampling distribution of a statistic by repeatedly resampling the sample data with replacement, treating it as if it were a fixed-size population~\citep{Chernick2011}. Although the idea of resampling was introduced in the 1930s by R. A. Fisher and E. J. G. Pitman, they performed sampling without replacement in those instances.

Bootstrap has already found applications in Natural Language Processing (NLP), particularly in analyzing the statistical significance of NLP systems. For example, \cite{Koehn2004} employed Bootstrap to estimate the statistical significance of the BLEU score in Machine Translation (MT). Similarly, \cite{Zhang2004} used Bootstrap to calculate confidence intervals for BLEU/NIST scores. In Automatic Speech Recognition (ASR), researchers have employed Bootstrap to estimate confidence intervals in performance evaluation, as evidenced in~\citep{Bisani2004}. Although Bootstrap in NLP is not a novel technique, it remains highly relevant and valuable.

\subsection{Comparison of Classifiers}
\label{sec:comparison1}


Comparing classification algorithms is a challenging and ongoing task. Numerous performance metrics have been introduced in classification, some specifically designed for comparing classifiers. In contrast, others were initially intended for different purposes. We aim to make inferences about the performance parameter $\theta$ of algorithms developed by participant systems in a challenge. These inferences are based on a single dataset of size $n$ and involve a minimal number of submissions. The goal is to estimate the parameter's value (performance) within the population from which the dataset is assumed to be randomly drawn.

Two conventional approaches for making parameter inferences are hypothesis testing and confidence intervals. In this contribution, performance inference employs both methods, specifically utilizing bootstrap estimates. The process involves generating $10,000$ Bootstrap samples with replacements from the dataset, each of the same size $n$ as the original gold standard examples and their corresponding predictions. Performance parameters for each team are computed for each sample, resulting in a sampling distribution. This distribution is used to construct a $95\%$ confidence interval for the mean of the performance parameter. Table \ref{tab:tabla-boot-media} provides the ordered estimates of the mean and corresponding confidence intervals obtained through the Bootstrap method, from the data presented in Table \ref{tab:metricas}.

\begin{table}[!h]
\caption{\label{tab:tabla-boot-media}Ordered Bootstrap Confidence Intervals}
\centering
\resizebox{\linewidth}{!}{
\fontsize{7}{9}\selectfont
\begin{tabular}[t]{lrrrlllll}
 \multicolumn{4}{c}{Basque}& & \multicolumn{4}{c}{Spanish}\\

  & LCI & mean & UCI & & & LCI & mean &UCI\\ 

\cellcolor{gray!6}{WordUp.01} & \cellcolor{gray!6}{0.5031} & \cellcolor{gray!6}{0.5716} & \cellcolor{gray!6}{0.6401}& & \cellcolor{gray!6}{WordUp.02} & \cellcolor{gray!6}{0.7734} & \cellcolor{gray!6}{0.8086} &\cellcolor{gray!6}{0.8437}\\ 

WordUp.02 & 0.4751 & 0.5444 & 0.6138& & WordUp.01 & 0.7537 & 0.7899 &0.8261\\ 

\cellcolor{gray!6}{MultiAztertest.01} & \cellcolor{gray!6}{0.4287} & \cellcolor{gray!6}{0.5007} & \cellcolor{gray!6}{0.5726}& & \cellcolor{gray!6}{MultiAztertest.01} & \cellcolor{gray!6}{0.6987} & \cellcolor{gray!6}{0.7400} &\cellcolor{gray!6}{0.7814}\\ 

SQYQP.01 & 0.3497 & 0.4237 & 0.4976& & SQYQP.01 & 0.6310 & 0.6730 &0.7149\\ 

\cellcolor{gray!6}{MultiAztertest.02} & \cellcolor{gray!6}{0.2664} & \cellcolor{gray!6}{0.3402} & \cellcolor{gray!6}{0.4139}& & \cellcolor{gray!6}{MultiAztertest.02} & \cellcolor{gray!6}{0.5945} & \cellcolor{gray!6}{0.6391} &\cellcolor{gray!6}{0.6837}\\ 

\end{tabular}}
\end{table}

\subsection{Comparison of Classifiers through Independents samples}
\label{sec:independents}

\begin{figure}[htbp]
\centering
\includegraphics[width=.5\textwidth]{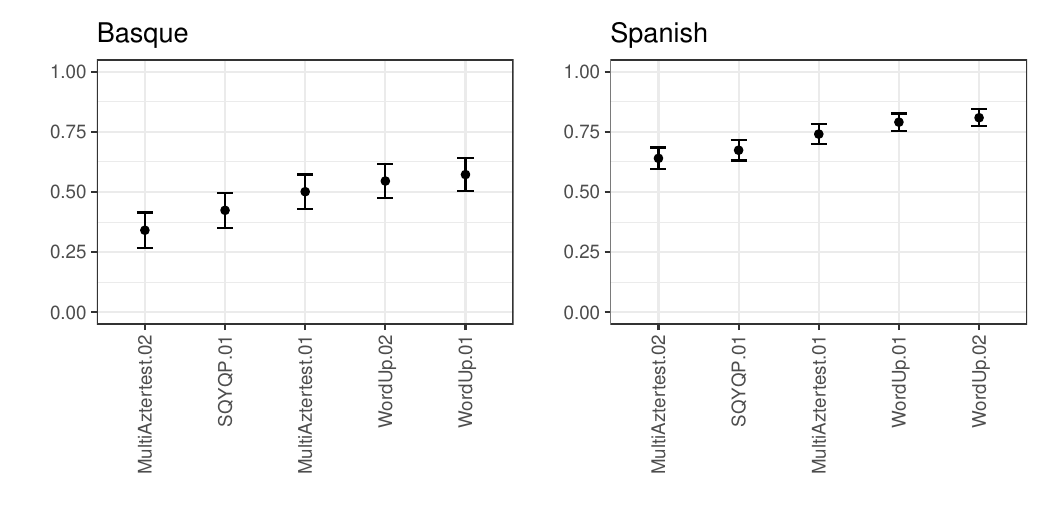}
\caption{Ordered Bootstrap Confidence Intervals}
\label{fig:grafica1}
\end{figure}

Suppose the confidence intervals of the means for two populations do not overlap. In that case, it indicates a significant difference in performance between the means of these populations. Conversely, if the intervals partially overlap, it indicates that there is some possibility that the means of the populations could be the same. The hypothesis testing approach involves setting the null hypothesis $H_0$ as $\theta_i = \theta_j$, and the alternative hypothesis $H_1$ as $\theta_i \neq \theta_j$, for $i \neq j$. Regarding performance metrics for the algorithms, one can observe the $95\%$ confidence intervals in Table \ref{tab:tabla-boot-media} and Figure \ref{fig:grafica1}. The intervals have been arranged for easier interpretation.

For Basque, the team with the highest score is $WordUp$ in run 1, with a $95\%$ confidence interval of $(0.5031, 0.6401)$. The second place goes to themselves in run 2 with an interval of $(0.4751, 0.6138)$. Since the first two intervals are very similar, the scores for both runs will likely be the same in the population from which the dataset was sampled. In contrast, there is a difference between $WordUp$ and $SQYQP$.

\subsection{Comparison of classifiers through paired samples}
\label{sec:paired}

\begin{figure}[htbp]
\centering
\includegraphics[width=.5\textwidth]{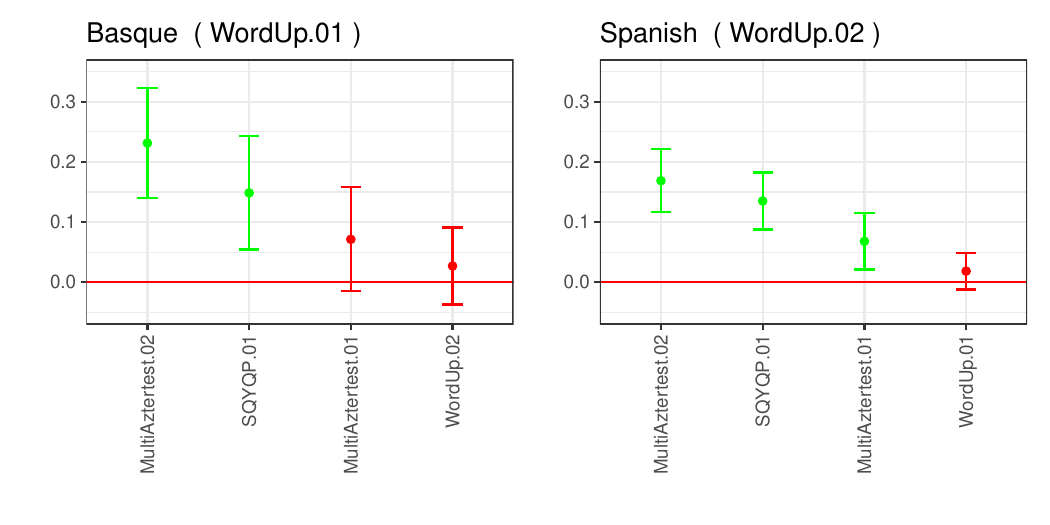}
\caption{Bootstrap Confidence Intervals of differences with the best. Red intervals contain zero, and green intervals do not contain it.}
\label{fig:grafica2}
\end{figure}

However, by including the gold standard and each team's prediction for the same element in each Bootstrap sample, we can calculate the performance and the performance difference for each sample. This approach is the paired bootstrap method \citep{Chernick2011}. Confidence intervals for the difference in performance between paired samples were constructed at the $95\%$ level, following the same methodology as in the previous case. Table \ref{tab:tabla-ref-media-2} and Figure \ref{fig:grafica2} present the confidence intervals for comparing the top-performing team with the others.

In this scenario, if the confidence interval for the performance difference contains zero, we cannot conclude that there is a significant difference in performance between the two algorithms in the population from which the dataset was drawn. In other words, we cannot reject the null hypothesis $H_0$. 

For the Basque dataset, the team with the highest performance is $WordUp$ in their first run. As observed, we cannot rule out that its performance is the same as their second run or $MultiAztertest$ with their first run. However, significant performance differences exist compared to the rest of the teams or runs. In the case of Spanish, a similar situation arises, with the best performance achieved by $WordUp$ in their second run, which is indistinguishable from their first run. Nevertheless, the rest of the teams and runs are different.

\begin{table}[h]

\caption{\label{tab:tabla-ref-media-2}Bootstrap Confidence Intervals of differences from the best.}
\centering
\resizebox{\linewidth}{!}{
\fontsize{7}{9}\selectfont
\begin{tabular}[t]{lrrrllrrr}
 \multicolumn{4}{c}{Basque (WordUp.01)}& & \multicolumn{4}{c}{Spanish (WordUp.02)}\\

  & LCI & mean & UCI & & & LCI & mean &UCI\\ 

\cellcolor{gray!6}{WordUp.02} & \cellcolor{gray!6}{-0.0371}& \cellcolor{gray!6}{0.0269}& \cellcolor{gray!6}{0.0910}& & \cellcolor{gray!6}{WordUp.01} & \cellcolor{gray!6}{-0.0120}& \cellcolor{gray!6}{0.0184}&\cellcolor{gray!6}{0.0488}\\

MultiAztertest.01 & -0.0152 & 0.0713& 0.1578& & MultiAztertest.01 & 0.0211& 0.0680&0.1149\\

\cellcolor{gray!6}{SQYQP.01} & \cellcolor{gray!6}{0.0543}& \cellcolor{gray!6}{0.1485}& \cellcolor{gray!6}{0.2427}& & \cellcolor{gray!6}{SQYQP.01} & \cellcolor{gray!6}{0.0877}& \cellcolor{gray!6}{0.1351}&\cellcolor{gray!6}{0.1825}\\

MultiAztertest.02 & 0.1405& 0.2314 & 0.3222& & MultiAztertest.02 & 0.1165& 0.1687&0.2210\\

\end{tabular}}
\end{table}

\subsection{Statistical Significance Testing}
\label{sec:significance}

In the previous sections, we displayed ordered confidence intervals for competitors' performance and compared them with the best submission; this prompts the question of whether we should test the hypothesis of equality versus difference, considering that one competitor outperforms the other in the test dataset.

To address this question, we compare the performance of two competitors, $A$ and $B$, to determine whether $A$ is superior to $B$ in a larger data population, represented as $\theta_A > \theta_B$. Suppose we have a test dataset $x = \{x_1, \ldots , x_n\}$, and let $A$ outperform $B$ by a margin of $\delta(x) = \theta_A (x) - \theta_B (x)$. The null hypothesis, denoted as $H_0$, posits that $A$ is not superior to $B$ in the overall population, while $H_1$ suggests the opposite. Therefore, we aim to determine the likelihood of observing a similar victory for $A$ in a new independent test dataset, denoted as $y$, assuming that $H_0$ holds.

Hypothesis testing aims to calculate the probability $p(\delta(X) > \delta(x) \mid H_0,x)$, where $X$ represents a random variable considering the possible test sets of size $n$ we could have chosen. In contrast, $\delta(x)$ denotes the observed difference, which is a constant. This probability is known as the $p-value(x)$. Traditionally, if $p-value(x) < 0.05$, the observed value $\delta(x)$ is considered sufficiently unlikely to reject $H_0$. In other words, the evidence suggests that $A$ is superior to $B$ \citep{Berg-Kirkpatrick2012}. 

Calculating the $p-value(x)$ is often complex and requires approximation methods to estimate it. This work employs the paired bootstrap method, not only because it is widely used \citep{Berg-Kirkpatrick2012,Bisani2004,Zhang2004,Koehn2004}, but also because it can be applied to any performance metric with ease.

\begin{figure}[htbp]
\centering
\includegraphics[width=.5\textwidth]{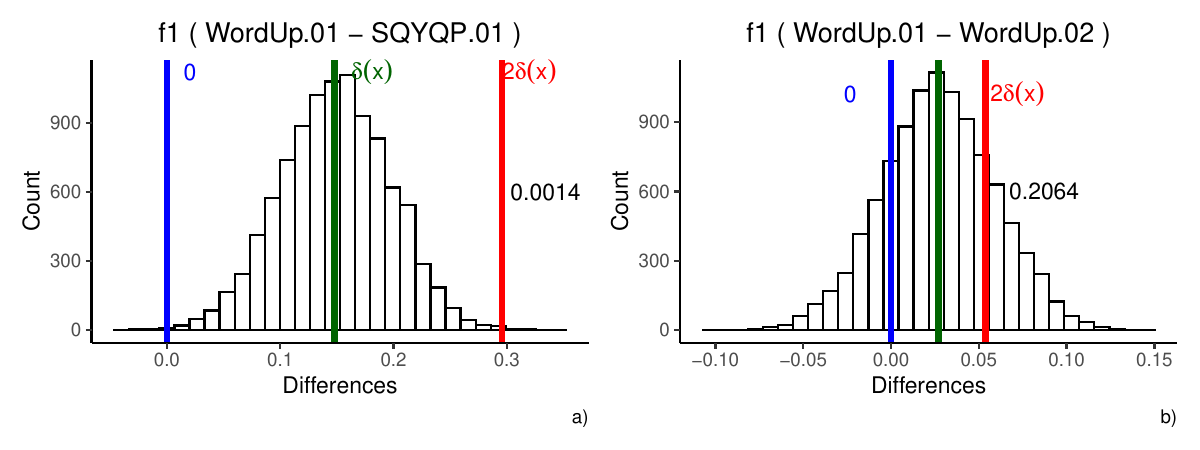}
\caption{ Bootstrap distribution of differences in the F1 macro-average score: (a) between $WordUp$ run 1 and $SQYQP$ run 1 for the Basque dataset, with an estimated $p-value$ of $0.0014$, and (b) between $WordUp$ run 1 and run 2 for the Basque dataset, with an estimated $p-value$ of $0.2064$.}
\label{fig:histogram}
\end{figure}

As demonstrated in \cite{Berg-Kirkpatrick2012}, we can estimate the $p-value(x)$ by calculating the fraction of instances where this difference exceeds $2\delta(x)$. It is crucial to note that this distribution is centered around $\delta(x)$ since $X$ is drawn from $x,$ where we observe that $A$ outperforms $B$ by $\delta(x)$. Figure \ref{fig:histogram} provides a visual representation of the $p-value(x)$ process, showcasing the bootstrap distribution of differences in the F1 macro-average score between $WordUp$ run 1 and $SQYQP$ run 1 (a) and between $WordUp$ run 1 and run 2 (b) for the Basque dataset. We highlight the values zero, $\delta(x)$, and $2\delta(x)$ for clarity.

When comparing $WordUp$ run 1 and $SQYQP$ run 1 in the test dataset $x,$ the difference $\delta(x) = 0.5734 - 0.4256 = 0.1478$ is statistically significant at the $5\%$ level because the $p-value(x)$ is $0.0014$. On the other hand, when comparing $WordUp$ run 1 and $WordUp$ run 2, $\delta(x) = 0.5734 - 0.5465 = 0.0269,$ which is not statistically significant at the $5\%$ level, with a $p-value(x)$ of $0.2064$. In other words, $WordUp$ run 1 is similar to $WordUp$ run 2 but is better than $SQYQP$ run 1.

\begin{table}[!h]
\caption{\label{tab:fscoretabeu} Differences of $F_1$ score (column)-(row), and their significance for Basque.  \\ Note: $\dagger p<.1$, {\footnotesize*}$p<.05$,
{\footnotesize**}$p<.01$,  and
{\footnotesize***}$p<.001$.}
\centering
\resizebox{\linewidth}{!}{
\fontsize{7}{9}\selectfont
\begin{tabular}[t]{lllll}
\toprule
  & WordUp.01 & WordUp.02 & MultiAztertest.01 & SQYQP.01\\
\midrule
\cellcolor{gray!6}{WordUp.02} & \cellcolor{gray!6}{0.027} & \cellcolor{gray!6}{ } & \cellcolor{gray!6}{ } & \cellcolor{gray!6}{ }\\
MultiAztertest.01 & 0.071 $\dagger$ & 0.044 &   &  \\
\cellcolor{gray!6}{SQYQP.01} & \cellcolor{gray!6}{0.148 ***} & \cellcolor{gray!6}{0.121 **} & \cellcolor{gray!6}{0.077 *} & \cellcolor{gray!6}{ }\\
MultiAztertest.02 & 0.231 *** & 0.204 *** & 0.160 *** & 0.083 *\\
\bottomrule
\end{tabular}}
\end{table}


\subsection{Multiple Comparisons}
\label{sec:MultipleComparison}

Multiple testing refers to any case involving simultaneous testing of multiple hypotheses. This situation is quite common in most empirical research. In our particular case, we are interested in conducting a multiple comparison of competitors' performance in a challenge. If we consider the ranking generated by ordering based on the performance metric chosen by the challenge organizers, our interest in terms of comparison is to find out whether the first place is better than the second, the third, ..., the $m-$th place, considering that the challenge has $m$ participants. If we denote $\theta_i(x)$ as the performance of the $i-$th place in the ranking, then the multiple hypotheses would be $H_0^j : \theta_1(x) \leq \theta_{j}(x) \; vs  \;   H_1^j : \theta_1(x) > \theta_{j}(x)$ with $j=2,3,\ldots,m.$ As explained in section \ref{sec:significance}, we obtain the corresponding p-value for each hypothesis test.

When multiple comparisons or hypothesis tests are performed on a dataset, the probability of making Type I errors (falsely rejecting a true null hypothesis) increases \citep{Jafari2019}; this is because conducting a larger number of tests raises the likelihood of finding statistically significant results by pure chance. It is essential to consider corrections for multiple comparisons for several reasons: (1) To control Type I Error, (2) To avoid spurious findings, and (3) To preserve research validity.

Now, we present a brief description of the most well-known methods for multiple comparison corrections, all of which adjust the p-values \citep{YoavBenjamini1995,Søgaard2014}: (1) Bonferroni correction: This method is conservative; it adjusts the significance level by dividing it by the number of comparisons. (2) Holm's step-down procedure: Holm's method adjusts the p-values sequentially. (3) Benjamini-Hochberg (BH) procedure: This procedure controls the false discovery rate (FDR), which is the expected proportion of false discoveries among all the rejected hypotheses.

In Table \ref{tab:fscoretabeu}, the observed performance differences, and their statistical significance are shown without any correction. Similar tables can be obtained by applying the multiple comparison corrections shown above. To illustrate the effects of these methods for multiple comparison corrections, Table \ref{tab:pvalues} is presented, applying them to the data in Table \ref{tab:fscoretabeu}. The next section will apply these multiple comparison fixes to compare metrics from different competitions.

\begin{table}[tbp]
\centering
\tiny
\caption{Estimated p-value for the F1 difference without adjustment and with Bonferroni, FDR, Holm, and BH adjusts}

\begin{tabular}{llrrrrrr}
\toprule
 &  & Difference & p-value & Bonferroni & FDR & Holm & BH\\
\midrule
WordUp.01 & WordUp.02 & 0.027 & 0.2030 & 0.8120 & 0.2030 & 0.2030 & 0.2030\\
WordUp.01 & MultiAztertest.01 & 0.071 & 0.0551 & 0.2204 & 0.0735 & 0.1102 & 0.0735\\
WordUp.01 & SQYQP.01 & 0.148 & 0.0012 & 0.0048 & 0.0024 & 0.0036 & 0.0024\\
WordUp.01 & MultiAztertest.02 & 0.231 & 0.0000 & 0.0000 & 0.0000 & 0.0000 & 0.0000\\
WordUp.02 & MultiAztertest.01 & 0.044 & 0.1490 & 0.4470 & 0.1490 & 0.1490 & 0.1490\\
WordUp.02 & SQYQP.01 & 0.121 & 0.0039 & 0.0117 & 0.0058 & 0.0078 & 0.0058\\
WordUp.02 & MultiAztertest.02 & 0.204 & 0.0000 & 0.0000 & 0.0000 & 0.0000 & 0.0000\\
MultiAztertest.01 & SQYQP.01 & 0.077 & 0.0330 & 0.0660 & 0.0330 & 0.0330 & 0.0330\\
MultiAztertest.01 & MultiAztertest.02 & 0.160 & 0.0003 & 0.0006 & 0.0006 & 0.0006 & 0.0006\\
SQYQP.01 & MultiAztertest.02 & 0.083 & 0.0427 & 0.0427 & 0.0427 & 0.0427 & 0.0427\\
\bottomrule
\end{tabular}
\label{tab:pvalues}
\end{table}

\section{Competitions Comparison}
\label{sec:CompetitionsComparison}

In Tables~\ref{tab:TASS2020} to~\ref{tab:Various}, we use the eight competitions as case studies to demonstrate the applicability of our methodology. Each table shows the information described in Table~\ref{tab:info-summary}.

\begin{table}[!h]
\centering
\scriptsize
\caption{Information Summary}
\begin{tabular}{>{\raggedright}p{2cm}>{\raggedright\arraybackslash}p{5cm}}
\hline 
Name & Description \\
\hline 
\cellcolor{gray!6}{$n$} & \cellcolor{gray!6}{Test data size} \\
$m$ & Number of participants or runs \\
\cellcolor{gray!6}{Ties w/ win} & \cellcolor{gray!6}{Possible ties with the winner (corrections: none/Bonferroni/Holm/BH)} \\
Poss. compars. & Total possible comparisons ($m \times (m-1)/2$) \\
\cellcolor{gray!6}{none/Bonf./Holm/BH} & \cellcolor{gray!6}{Ties between competitors with corrections (none/Bonferroni/Holm/BH)} \\
$|win.-med|$ & Performance difference between the winner and the competitor in the middle of the table \\
\cellcolor{gray!6}{CV} & \cellcolor{gray!6}{Coefficient of variation of competitors' performance. ($CV=100\times s_x/\overline{x}$, where $\overline{x}$ is the mean of $x$ and $s_x$ is the standard deviation of $x$)} \\
PPI & Possible Percentage Improvement, e.g., for F1 score, it's calculated as $100\times(1-F_1^{winner})$ \\
\hline
\end{tabular}
\label{tab:info-summary}
\end{table}

\begin{table}[htbp]
\centering
\scriptsize
\caption{Results for the TASS2020 challenge.}

\begin{tabular}{|l|rrrrr|}
\hline
Task           & \multicolumn{5}{c|}{General polarity at three levels}                                                                                                      \\ \hline
Language       & \multicolumn{1}{c|}{Spain}   & \multicolumn{1}{c|}{Peru}    & \multicolumn{1}{c|}{Costa Rica} & \multicolumn{1}{c|}{Uruguay} & \multicolumn{1}{c|}{Mexico} \\ \hline
Metric         & \multicolumn{5}{c|}{macro-averaged F1 score}                                                                                                               \\ \hline
$n$            & \multicolumn{1}{r|}{1706}    & \multicolumn{1}{r|}{1464}    & \multicolumn{1}{r|}{1166}       & \multicolumn{1}{r|}{1428}    & 1500                        \\ \hline
$m$            & \multicolumn{1}{r|}{3}       & \multicolumn{1}{r|}{3}       & \multicolumn{1}{r|}{3}          & \multicolumn{1}{r|}{3}       & 3                           \\ \hline
Ties w/ win.   & \multicolumn{1}{r|}{1/1/1/1} & \multicolumn{1}{r|}{1/1/1/1} & \multicolumn{1}{r|}{1/1/1/1}    & \multicolumn{1}{r|}{1/1/1/1} & 1/1/1/1                     \\ \hline
Poss. compars. & \multicolumn{1}{r|}{3}       & \multicolumn{1}{r|}{3}       & \multicolumn{1}{r|}{3}          & \multicolumn{1}{r|}{3}       & 3                           \\ \hline
none/Bonf.     & \multicolumn{1}{r|}{1/1}     & \multicolumn{1}{r|}{1/1}     & \multicolumn{1}{r|}{1/1}        & \multicolumn{1}{r|}{1/1}     & 1/1                         \\ \hline
Holm/BH        & \multicolumn{1}{r|}{1/1}     & \multicolumn{1}{r|}{1/1}     & \multicolumn{1}{r|}{1/1}        & \multicolumn{1}{r|}{1/1}     & 1/1                         \\ \hline
$|win.-med|$   & \multicolumn{1}{r|}{0.010}    & \multicolumn{1}{r|}{0.008}    & \multicolumn{1}{r|}{0.001}       & \multicolumn{1}{r|}{0.016}    & 0.002                        \\ \hline
$CV$             & \multicolumn{1}{r|}{24.010}  & \multicolumn{1}{r|}{27.310}  & \multicolumn{1}{r|}{31.234}     & \multicolumn{1}{r|}{12.910}  & 24.625                      \\ \hline
PPI            & \multicolumn{1}{r|}{32.98}   & \multicolumn{1}{r|}{36.647}  & \multicolumn{1}{r|}{35.365}     & \multicolumn{1}{r|}{33.669}  & 36.599                      \\ \hline
\end{tabular}

\label{tab:TASS2020}
\end{table}

There are different aspects to consider when comparing different NLP competitions. The first aspect is to ask whether the winner is better than the other competitors or runs. In this regard, we can see that in all competitions, there is at least one tie with the winner, except in the Close Track-Contextual in Basque in the VaxxStance competition, in the Sentiment task of REST-MEX 2021, and even in DETOXIS.

\begin{table}[htbp]
\centering
\scriptsize
\caption{Results for the VaxxStance challenge.}
\begin{tabular}{|l|rrrr|}
\hline
Task           & \multicolumn{2}{c|}{Close Track-Textual}                    & \multicolumn{2}{c|}{Close Track-Contextual}                \\ \hline
Language       & \multicolumn{1}{c|}{Spanish} & \multicolumn{1}{c|}{Basque}  & \multicolumn{1}{c|}{Spanish} & \multicolumn{1}{c|}{Basque} \\ \hline
Metric         & \multicolumn{4}{c|}{macro-averaged F1 score(FAVOR, AGAINST).}                                                            \\ \hline
$n$            & \multicolumn{1}{r|}{694}     & \multicolumn{1}{r|}{312}     & \multicolumn{1}{r|}{694}     & 312                         \\ \hline
$m$            & \multicolumn{1}{r|}{5}       & \multicolumn{1}{r|}{5}       & \multicolumn{1}{r|}{5}       & 5                           \\ \hline
Ties W/ Win.   & \multicolumn{1}{r|}{1/1/1/1} & \multicolumn{1}{r|}{2/2/2/2} & \multicolumn{1}{r|}{1/1/1/1} & 0/0/0/0                     \\ \hline
Poss. Compars. & \multicolumn{1}{r|}{10}      & \multicolumn{1}{r|}{10}      & \multicolumn{1}{r|}{10}      & 10                          \\ \hline
None/Bonf.     & \multicolumn{1}{r|}{2/2}     & \multicolumn{1}{r|}{3/4}     & \multicolumn{1}{r|}{2/2}     & 1/1                         \\ \hline
Holm/BH        & \multicolumn{1}{r|}{2/2}     & \multicolumn{1}{r|}{3/3}     & \multicolumn{1}{r|}{2/2}     & 1/1                         \\ \hline
$|win.-med|$   & \multicolumn{1}{r|}{0.068}    & \multicolumn{1}{r|}{0.071}    & \multicolumn{1}{r|}{0.098}    & 0.410                        \\ \hline
$CV$           & \multicolumn{1}{r|}{9.970}    & \multicolumn{1}{r|}{19.680}   & \multicolumn{1}{r|}{10.463}  & 64.766                      \\ \hline
PPI            & \multicolumn{1}{r|}{19.084}  & \multicolumn{1}{r|}{42.660}  & \multicolumn{1}{r|}{10.871}  & 22.291                      \\ \hline
\end{tabular}
\label{tab:VaxxStance}
\end{table}

The competitiveness of a competition is evaluated with three aspects; the first is the number of ties in relation to the possible comparisons. When there are many potential ties, it indicates a higher level of competitiveness. The second aspect analyzed is the coefficient of variation ($CV$) of participants' performance. A smaller $CV$ means a more significant competitiveness. The third aspect is the distance between the winner and the middle competitor ($|win. - med.|$). With this measure, a smaller distance indicates a smaller gap.

It can be observed from the tables that the most competitive task is Sexism Identification for the English language, as it has the smallest $CV$ with a value of $0.78\%$. It also shows that almost all comparisons result in ties compared to the winner. Additionally, it has one of the smallest $|win. - med.|$ values, which is $0.78$. It is worth noting that this task analyzed only the top 10 participants. Something similar happens with the other EXIST subtasks involving only one language (English, Spanish). If we do not consider these cases, one of the most competitive competitions is PAR-MEX 2022, where slightly less than a quarter of the total comparisons result in ties. It has a $CV$ of $4.72\%$ and a $|win. - med.|$ of $0.061.$ At the other extreme, we can find tasks like Close Track - Contextual in Basque from VaxxStance with a $CV$ of $64.76\%$ and a $|win. - med.|$ of $0.410$. Furthermore, one out of every ten comparisons resulted in a tie. Due to the nature of the MAE metric, this aspect does not include the analysis of its $CV$.

The calculation of Possible Percentage Improvement (PPI) is proposed to assess the potential of a task considering its metric. This indicator will be higher when the gap between the performance of the so-called winner is large compared to the ideal value of the performance metric. This indicator works for scores where the highest value is the best, and the scores are capped, which in these competitions would be all metrics except REST-MEX 2021 with the MAE metric. When competition has a low $PPI$ value, achieving substantial improvements will be more challenging. The competitions that were found to have the highest potential for improvement are MEX-A3T 2019, REST-MEX 2022 in the Epidemiological Semaphore task (although this task was a particular case due to the pandemic), EXIST in the Sexism Categorization task, and VaxxStance in the Close Track - Textual task in Basque. All of these tasks and competitions had values exceeding $39\%$.

\begin{table}[htbp]
\centering
\tiny
\caption{Results for the EXIST Challenge.}
\begin{tabular}{|l|rrr|rrr|}
\hline
Task           & \multicolumn{3}{c|}{Sexism Identification}                                                 & \multicolumn{3}{c|}{Sexism Categorization}                                                 \\ \hline
Language       & \multicolumn{1}{c|}{All}     & \multicolumn{1}{c|}{English} & \multicolumn{1}{c|}{Spanish} & \multicolumn{1}{c|}{All}     & \multicolumn{1}{c|}{English} & \multicolumn{1}{c|}{Spanish} \\ \hline
Metric         & \multicolumn{3}{c|}{accuracy}                                                              & \multicolumn{3}{c|}{macro-averaged F1 score}                                               \\ \hline
$n$            & \multicolumn{1}{r|}{4368}    & \multicolumn{1}{r|}{2208}    & 2160                         & \multicolumn{1}{r|}{4368}    & \multicolumn{1}{r|}{2208}    & 2160                         \\ \hline
$m$            & \multicolumn{1}{r|}{31}      & \multicolumn{1}{r|}{10}      & 10                           & \multicolumn{1}{r|}{28}      & \multicolumn{1}{r|}{10}      & 10                           \\ \hline
Ties w/ win.   & \multicolumn{1}{r|}{4/9/7/5} & \multicolumn{1}{r|}{5/9/9/9} & 3/7/7/4                      & \multicolumn{1}{r|}{2/6/2/2} & \multicolumn{1}{r|}{4/7/7/4} & 2/5/4/2                      \\ \hline
Poss. compars. & \multicolumn{1}{r|}{465}     & \multicolumn{1}{r|}{45}      & 45                           & \multicolumn{1}{r|}{378}     & \multicolumn{1}{r|}{45}      & 45                           \\ \hline
none/Bonf.     & \multicolumn{1}{r|}{81/133}  & \multicolumn{1}{r|}{41/45}   & 28/41                        & \multicolumn{1}{r|}{62/101}  & \multicolumn{1}{r|}{35/43}   & 31/40                        \\ \hline
Holm/BH        & \multicolumn{1}{r|}{118/89}  & \multicolumn{1}{r|}{45/45}   & 41/37                        & \multicolumn{1}{r|}{82/68}   & \multicolumn{1}{r|}{43/36}   & 39/33                        \\ \hline
$|win.-med|$   & \multicolumn{1}{r|}{0.029}   & \multicolumn{1}{r|}{0.011}   & 0.016                        & \multicolumn{1}{r|}{0.053}    & \multicolumn{1}{r|}{0.021}   & 0.029                         \\ \hline
$CV$           & \multicolumn{1}{r|}{10.920}   & \multicolumn{1}{r|}{0.780}    & 1.190                         & \multicolumn{1}{r|}{24.120}   & \multicolumn{1}{r|}{2.140}    & 2.260                         \\ \hline
PPI            & \multicolumn{1}{r|}{21.95}   & \multicolumn{1}{r|}{22.28}   & 20.55                        & \multicolumn{1}{r|}{42.13}   & \multicolumn{1}{r|}{43.96}   & 39.27                        \\ \hline
\end{tabular}
\label{tab:EXIST}
\end{table}

\begin{table}[htbp]
\centering
\scriptsize
\caption{Results for the REST-MEX Challenge.}
\begin{tabular}{|l|rr|rr|}
\hline
Challenge      & \multicolumn{2}{c|}{REST-MEX 2021}                                   & \multicolumn{2}{c|}{REST-MEX 2022}               \\ \hline
Task           & \multicolumn{1}{c|}{Recommendation} & \multicolumn{1}{c|}{Sentiment} & \multicolumn{1}{c|}{Sentiment}   & Epi Semaphore \\ \hline
Metric         & \multicolumn{2}{c|}{MAE}                                             & \multicolumn{1}{r|}{$measure_S$} & $measure_C$   \\ \hline
$n$            & \multicolumn{1}{r|}{681}            & 2216                           & \multicolumn{1}{r|}{12938}       & 744           \\ \hline
$m$            & \multicolumn{1}{r|}{4}              & 15                             & \multicolumn{1}{r|}{27}          & 15            \\ \hline
Ties w/ win.   & \multicolumn{1}{r|}{1/1/1/1}        & 0/0/0/0                        & \multicolumn{1}{r|}{2/4/2/2}     & 1/1/1/1       \\ \hline
Poss. compars. & \multicolumn{1}{r|}{6}              & 105                            & \multicolumn{1}{r|}{351}         & 105           \\ \hline
none/Bonf.     & \multicolumn{1}{r|}{1/1}            & 8/12                           & \multicolumn{1}{r|}{16/31}       & 8/14          \\ \hline
Holm/BH        & \multicolumn{1}{r|}{1/1}            & 9/8                            & \multicolumn{1}{r|}{18/16}       & 8/8           \\ \hline
$|win.-med|$   & \multicolumn{1}{r|}{0.212}           & 0.193                           & \multicolumn{1}{r|}{0.023}        & 0.161          \\ \hline
$CV$           & \multicolumn{1}{r|}{84.283}         & 28.740                         & \multicolumn{1}{r|}{14.370}      & 38.557        \\ \hline
PPI            & \multicolumn{1}{r|}{0.310}          & .475                           & \multicolumn{1}{r|}{10.761}      & 51.001        \\ \hline
\end{tabular}
\label{tab:REST-MEX}
\end{table}

\begin{table}[htbp]
\centering
\tiny
\caption{Results for the Various Challenges.}
\begin{tabular}{|l|rrrrr|}
\hline
\multirow{2}{*}{Challenge} & \multicolumn{1}{c|}{DETOXIS}   & \multicolumn{1}{c|}{PAR-MEX}        & \multicolumn{1}{c|}{MeOffendEs} & \multicolumn{2}{c|}{MEX-A3T}                                       \\
                        & \multicolumn{1}{c|}{2021}      & \multicolumn{1}{c|}{2022}           & \multicolumn{1}{c|}{2021}       & \multicolumn{2}{c|}{2019}                                          \\ \hline
\multirow{2}{*}{Task}   & \multicolumn{1}{c|}{Toxicity}  & \multicolumn{1}{c|}{Paraphrase}     & \multicolumn{1}{c|}{Non}        & \multicolumn{1}{c|}{Agg}     & \multicolumn{1}{c|}{author}         \\
                        & \multicolumn{1}{c|}{detection} & \multicolumn{1}{c|}{Identification} & \multicolumn{1}{c|}{contextual} & \multicolumn{1}{l|}{}        & \multicolumn{1}{c|}{profiling}      \\ \hline
\multirow{2}{*}{Metric} & \multicolumn{4}{c|}{F1 score}                                                                                                         & \multicolumn{1}{c|}{macro-averaged} \\
                        & \multicolumn{4}{l|}{}                                                                                                                 & \multicolumn{1}{c|}{F1 score}       \\ \hline
$n$                     & \multicolumn{1}{r|}{891}       & \multicolumn{1}{r|}{2821}           & \multicolumn{1}{r|}{2182}       & \multicolumn{1}{r|}{3156}    & 1500                                \\ \hline
$m$                     & \multicolumn{1}{r|}{31}        & \multicolumn{1}{r|}{8}              & \multicolumn{1}{r|}{10}         & \multicolumn{1}{r|}{25}      & 4                                   \\ \hline
Ties w/ win.            & \multicolumn{1}{r|}{0/3/0/0}   & \multicolumn{1}{r|}{1/1/1/1}        & \multicolumn{1}{r|}{1/2/2/1}    & \multicolumn{1}{r|}{3/7/4/3} & 1/1/1/1                             \\ \hline
Poss. compars.          & \multicolumn{1}{r|}{465}       & \multicolumn{1}{r|}{28}             & \multicolumn{1}{r|}{45}         & \multicolumn{1}{r|}{300}     & 6                                   \\ \hline
none/Bonf.              & \multicolumn{1}{r|}{80/135}    & \multicolumn{1}{r|}{6/6}            & \multicolumn{1}{r|}{7/9}        & \multicolumn{1}{r|}{70/91}   & 2/2                                 \\ \hline
Holm/BH                 & \multicolumn{1}{r|}{112/85}    & \multicolumn{1}{r|}{6/6}            & \multicolumn{1}{r|}{8/7}        & \multicolumn{1}{r|}{80/63}   & 2/2                                 \\ \hline
$|win.-med|$            & \multicolumn{1}{r|}{0.223}     & \multicolumn{1}{r|}{0.061}          & \multicolumn{1}{r|}{0.078}      & \multicolumn{1}{r|}{0.098}   & 0.164                               \\ \hline
$CV$                    & \multicolumn{1}{r|}{42.600}    & \multicolumn{1}{r|}{4.722}          & \multicolumn{1}{r|}{16.070}     & \multicolumn{1}{r|}{19.620}  & 46.491                              \\ \hline
PPI                     & \multicolumn{1}{r|}{35.390}    & \multicolumn{1}{r|}{5.758}          & \multicolumn{1}{r|}{28.46}      & \multicolumn{1}{r|}{52.038}  & 42.581                              \\ \hline
\end{tabular}
\label{tab:Various}
\end{table}

\section{Conclusions}
\label{sec:conclusions}

This research proposes a general-purpose methodology that employs bootstrapping to estimate the performance of different teams in a competition. This estimation can be performed using various metrics employed in the competition or solely relying on the metric used for ranking competitors. It involves the construction of confidence intervals for each competitor and for performance differences when compared to the winner, all based on the bootstrap method. Visualizing these confidence intervals provides a handy tool for quickly assessing whether observed differences hold statistical significance or are merely a result of randomness. Additionally, significance calculations were presented to determine if one competitor significantly outperforms another. 

Furthermore, we presented the construction of a lower diagonal matrix with comparisons between competitors, displaying the difference and the statistical significance of this difference. It was emphasized that using adjustment methods in multiple comparisons is essential to ensure the statistical integrity of the results and make decisions based on solid evidence in challenges and competitions. Our methodology can be easily applied to any classification or regression problem. Additionally, the analysis covered a range of NLP competitions that differed in class numbers, metric usage, test set sizes, and the number of competitors involved. We have compared these challenges in terms of their competitiveness, assessing their potential for improvement in future editions. We also analyzed whether the so-called winner's advantage was clear.

\textbf{Acknowledgments.} Authors are extremely grateful with organizers of IberLEF competitions for sharing the results of their respective tasks. 

\bibliographystyle{apalike}
\bibliography{mybibliography}

\begin{thebibliography}{}

\bibitem[Agerri et~al., 2021]{Agerri2021}
Agerri, R., Centeno, R., Espinosa, M., de~Landa, J.~F., and Rodrigo, {\'{A}}.
  (2021).
\newblock {VaxxStance@IberLEF 2021: Overview of the Task on Going Beyond Text
  in Cross-Lingual Stance Detection}.
\newblock {\em Procesamiento del Lenguaje Natural}, 67(0):173--181.

\bibitem[{\'{A}}lvarez-Carmona et~al., 2021]{Alvarez-Carmona2021}
{\'{A}}lvarez-Carmona, M.~{\'{A}}., Aranda, R., Arce-Cardenas, S.,
  Fajardo-Delgado, D., Guerrero-Rodr{\'{i}}guez, R., L{\'{o}}pez-Monroy, A.~P.,
  Mart{\'{i}}nez-Miranda, J., P{\'{e}}rez-Espinosa, H., and
  Rodr{\'{i}}guez-Gonz{\'{a}}lez, A.~Y. (2021).
\newblock {Overview of Rest-Mex at IberLEF 2021: Recommendation System for Text
  Mexican Tourism}.
\newblock {\em Procesamiento del Lenguaje Natural}, 67(0):163--172.

\bibitem[{\'{A}}lvarez-Carmona et~al., 2022]{Alvarez-Carmona2022}
{\'{A}}lvarez-Carmona, M.~{\'{A}}., D{\'{i}}az-Pacheco, {\'{A}}., Aranda, R.,
  Rodr{\'{i}}guez-Gonz{\'{a}}lez, A.~Y., Fajardo-Delgado, D.,
  Guerrero-Rodr{\'{i}}guez, R., and Bustio-Mart{\'{i}}nez, L. (2022).
\newblock {Overview of Rest-Mex at IberLEF 2022: Recommendation System,
  Sentiment Analysis and Covid Semaphore Prediction for Mexican Tourist Texts}.
\newblock {\em Procesamiento del Lenguaje Natural}, 69(0):289--299.

\bibitem[Arag{\'{o}}n et~al., 2019]{Aragon2019}
Arag{\'{o}}n, M.~E., {\'{A}}lvarez-Carmona, M., Montes-Y-G{\'{o}}mez, M.,
  Escalante, H.~J., Villase{\~{n}}or-Pineda, L., and Moctezuma, D. (2019).
\newblock {Overview of MEX-A3T at IberLEF 2019: Authorship and aggressiveness
  analysis in Mexican Spanish tweets}.
\newblock {\em CEUR Workshop Proceedings}, 2421:478--494.

\bibitem[Bel-Enguix et~al., 2022]{Bel-Enguix2022}
Bel-Enguix, G., Sierra, G., G{\'{o}}mez-Adorno, H., Torres-Moreno, J.-M.,
  Ortiz-Barajas, J.-G., and V{\'{a}}squez, J. (2022).
\newblock {Overview of PAR-MEX at Iberlef 2022: Paraphrase Detection in Spanish
  Shared Task}.
\newblock {\em Procesamiento del Lenguaje Natural}, 69(0):255--263.

\bibitem[Berg-Kirkpatrick et~al., 2012]{Berg-Kirkpatrick2012}
Berg-Kirkpatrick, T., Burkett, D., and Klein, D. (2012).
\newblock {An empirical investigation of statistical significance in NLP}.
\newblock In {\em EMNLP-CoNLL 2012 - 2012 Joint Conference on Empirical Methods
  in Natural Language Processing and Computational Natural Language Learning,
  Proceedings of the Conference}.

\bibitem[Bisani and Ney, 2004]{Bisani2004}
Bisani, M. and Ney, H. (2004).
\newblock {Bootstrap estimates for confidence intervals in ASR performance
  evaluation}.
\newblock {\em 2004 IEEE International Conference on Acoustics, Speech, and
  Signal Processing}, 1.

\bibitem[Chernick and LaBudde, 2011]{Chernick2011}
Chernick, M.~R. and LaBudde, R.~A. (2011).
\newblock {\em {An introduction to bootstrap methods with applications to R}}.
\newblock Wiley.

\bibitem[Dem{\v{s}}ar, 2006]{JMLR:v7:demsar06a}
Dem{\v{s}}ar, J. (2006).
\newblock Statistical comparisons of classifiers over multiple data sets.
\newblock {\em Journal of Machine Learning Research}, 7(1):1--30.

\bibitem[Dietterich, 1998]{Dietterich1998}
Dietterich, T.~G. (1998).
\newblock {Approximate Statistical Tests for Comparing Supervised
  Classification Learning Algorithms}.
\newblock {\em Neural Computation}, 10(7):1895--1923.

\bibitem[Everingham et~al., 2010]{voc}
Everingham, M., Gool, L.~V., Williams, C. K.~I., Winn, J.~M., and Zisserman, A.
  (2010).
\newblock The pascal visual object classes {(VOC)} challenge.
\newblock {\em Int. J. Comput. Vis.}, 88(2):303--338.

\bibitem[Garc{\'{i}}a and Herrera, 2008]{Garcia2008}
Garc{\'{i}}a, S. and Herrera, F. (2008).
\newblock {An extension on ``statistical comparisons of classifiers over
  multiple data sets'' for all pairwise comparisons}.
\newblock {\em Journal of Machine Learning Research}, 9:2677--2694.

\bibitem[Garc{\'{i}a}-Vega et~al., 2020]{Garcia-Vegaa2020}
Garc{\'{i}a}-Vega, M., D{\'{i}}az-Galiano, M.~C., Garc{\'{i}}a-Cumbreras, M.,
  {Del Arco}, F. M.~P., Montejo-R{\'{a}}ez, A., Jim{\'{e}}nez-Zafra, S.~M.,
  C{\'{a}}mara, E.~M., Aguilar, C.~A., Cabezudo, M. A.~S., Chiruzzo, L., and
  Moctezuma, D. (2020).
\newblock {Overview of TASS 2020: Introducing Emotion Detection}.
\newblock {\em CEUR Workshop Proceedings}, 2664:163--170.

\bibitem[Jafari and Ansari-Pour, 2019]{Jafari2019}
Jafari, M. and Ansari-Pour, N. (2019).
\newblock {Why, When and How to Adjust Your P Values?}
\newblock {\em Cell Journal (Yakhteh)}, 20(4):604.

\bibitem[Koehn, 2004]{Koehn2004}
Koehn, P. (2004).
\newblock {Statistical Significance Tests for Machine Translation Evaluation}.
\newblock In {\em Proceedings of the 2004 Conference on Empirical Methods in
  Natural Language Processing, EMNLP 2004 - A meeting of SIGDAT, a Special
  Interest Group of the ACL held in conjunction with ACL 2004}, pages 388--395.

\bibitem[Nava-Mu{\~{n}}oz et~al., 2023]{10.1007/978-3-031-33783-3_9}
Nava-Mu{\~{n}}oz, S., Graff~Guerrero, M., and Escalante, H.~J. (2023).
\newblock Comparison of classifiers in challenge scheme.
\newblock In Rodr{\'i}guez-Gonz{\'a}lez, A.~Y., P{\'e}rez-Espinosa, H.,
  Mart{\'i}nez-Trinidad, J.~F., Carrasco-Ochoa, J.~A., and Olvera-L{\'o}pez,
  J.~A., editors, {\em Pattern Recognition}, pages 89--98, Cham. Springer
  Nature Switzerland.

\bibitem[Pavao et~al., 2023]{JMLR:v24:21-1436}
Pavao, A., Guyon, I., Letournel, A.-C., Tran, D.-T., Baro, X., Escalante,
  H.~J., Escalera, S., Thomas, T., and Xu, Z. (2023).
\newblock Codalab competitions: An open source platform to organize scientific
  challenges.
\newblock {\em Journal of Machine Learning Research}, 24(198):1--6.

\bibitem[{Plaza-del-Arco} et~al., 2021]{Plaza-del-Arco2021}
{Plaza-del-Arco}, F.~M., Casavantes, M., Escalante, H.~J.,
  Mart{\'{i}}n-Valdivia, M.~T., Montejo-R{\'{a}}ez, A., Montes-y G{\'{o}}mez,
  M., Jarqu{\'{i}}n-V{\'{a}}squez, H., and Villase{\~{n}}or-Pineda, L. (2021).
\newblock {Overview of MeOffendEs at IberLEF 2021: Offensive Language Detection
  in Spanish Variants}.
\newblock {\em Procesamiento del Lenguaje Natural}, 67(0):183--194.

\bibitem[Rodr{\'{i}}guez-S{\'{a}}nchez et~al., 2021]{Rodriguez-Sanchez2021}
Rodr{\'{i}}guez-S{\'{a}}nchez, F., Carrillo-de Albornoz, J., Plaza, L.,
  Gonzalo, J., Rosso, P., Comet, M., and Donoso, T. (2021).
\newblock {Overview of EXIST 2021: sEXism Identification in Social neTworks}.
\newblock {\em Procesamiento del Lenguaje Natural}, 67(0):195--207.

\bibitem[Russakovsky et~al., 2015]{imagenet}
Russakovsky, O., Deng, J., Su, H., Krause, J., Satheesh, S., Ma, S., Huang, Z.,
  Karpathy, A., Khosla, A., Bernstein, M.~S., Berg, A.~C., and Fei{-}Fei, L.
  (2015).
\newblock Imagenet large scale visual recognition challenge.
\newblock {\em Int. J. Comput. Vis.}, 115(3):211--252.

\bibitem[S{\o}gaard et~al., 2014]{Søgaard2014}
S{\o}gaard, A., Johannsen, A., Plank, B., Hovy, D., and Martinez, H. (2014).
\newblock {What's in a p-value in NLP?}
\newblock {\em CoNLL 2014 - 18th Conference on Computational Natural Language
  Learning, Proceedings}, pages 1--10.

\bibitem[Taul{\'{e}} et~al., 2021]{Taule2021}
Taul{\'{e}}, M., Ariza, A., Nofre, M., Amig{\'{o}}, E., and Rosso, P. (2021).
\newblock {Overview of DETOXIS at IberLEF 2021: DEtection of TOXicity in
  comments In Spanish}.
\newblock {\em Procesamiento del Lenguaje Natural}, 67(0):209--221.

\bibitem[{Yoav Benjamini} and {Yosef Hochberg}, 1995]{YoavBenjamini1995}
{Yoav Benjamini} and {Yosef Hochberg} (1995).
\newblock {Controlling the False Discovery Rate: A Practical and Powerful
  Approach to Multiple Testing}.
\newblock {\em Journal of the Royal Statistical Society. Series B
  (Methodological)}, 57(1):289--300.

\bibitem[Zhang et~al., 2004]{Zhang2004}
Zhang, Y., Vogel, S., and Waibel, A. (2004).
\newblock {Interpreting BLEU/NIST Scores: How Much Improvement do We Need to
  Have a Better System?}
\newblock In {\em Proceedings of the 4th International Conference on Language
  Resources and Evaluation, LREC 2004}, pages 2051--2054.

\end{thebibliography}

\appendix
\section{CompStats Package}

This appendix illustrates the package {\em CompStats} that implements the ideas presented in this contribution. The first step is to install the library, which can be done using {\em pip} as follows: 

\begin{lstlisting}[language=Bash]
pip install CompStats
\end{lstlisting}

Once CompStats is installed, one must load a few libraries.

\begin{lstlisting}[language=Python]
from CompStats import performance, difference, plot_difference
from statsmodels.stats.multitest import multipletests
from sklearn.metrics import f1_score
import pandas as pd
\end{lstlisting}
To illustrate CompStats, we will use the PAR-MEX 2022 dataset. Let us assume \emph{PARMEX\_2022.csv} is a csv file where the column $y$ has the ground truth, and the other columns are the systems' outputs.
\begin{lstlisting}[language=Python]
DATA = "PARMEX_2022.csv"
df = pd.read_csv(DATA)
\end{lstlisting}

The performance metric used is the F1 score.
\begin{lstlisting}[language=Python]
score = lambda y, hy:f1_score(y, hy)
\end{lstlisting}

The following instructions calculate the individual performance of each system,  the difference in their performance relative to the best-performing system, and compare the analyzed algorithms through confidence intervals. 
\begin{lstlisting}[language=Python]
perf = performance(df, score=score)
diff = difference(perf)
ins = plot_difference(diff)
\end{lstlisting}

\begin{figure}[htbp]
\centering
\includegraphics[width=.3\textwidth]{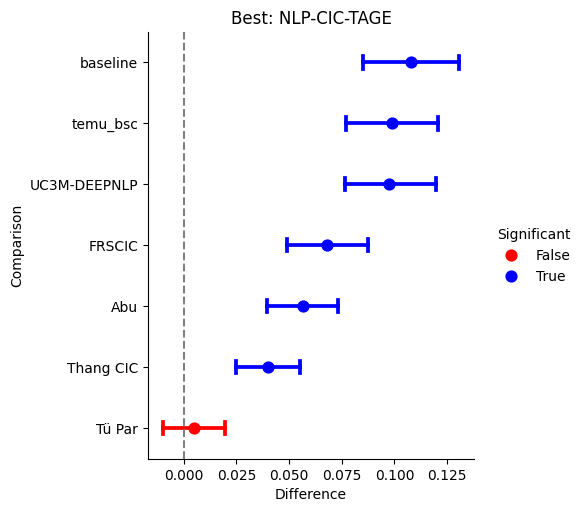}
\caption{Ordered Bootstrap Confidence Intervals for performance differences.}
\label{fig:CompStatsdifferences}
\end{figure}

A p-value is calculated to determine if performance differences are statistically significant, offering a concise way to assess the reliability of the observed differences.

\begin{lstlisting}[language=Python]
p_values = difference_p_value(diff)
p_values
\end{lstlisting}
\begin{verbatim}
{'baseline': 0.0, 'temu_bsc': 0.0,
 'UC3M-DEEPNLP': 0.0, 'FRSCIC': 0.0,
 'Abu': 0.0, 'Thang CIC': 0.0,
 'Tü Par': 0.254}
\end{verbatim} 

Finally, a Bonferroni correction is applied to the p-values, exemplifying this method to ensure a rigorous evaluation of significance across multiple tests.

\begin{lstlisting}[language=Python]
result = multipletests(list(p_values.values()), method='bonferroni')
p_valuesC = dict(zip(p_values.keys(),result[1]))
p_valuesC
\end{lstlisting}
\begin{verbatim}
{'baseline': 0.0, 'temu_bsc': 0.0,
 'UC3M-DEEPNLP': 0.0, 'FRSCIC': 0.0,
 'Abu': 0.0, 'Thang CIC': 0.0,
 'Tü Par': 1.0}
\end{verbatim} 

\end{document}